\pdfoutput=1
\documentclass[11pt]{article}

\usepackage{EACL2023}

\usepackage{latexsym}

\usepackage[T2A,T1]{fontenc}
\usepackage[utf8]{inputenc}

\usepackage{microtype}

\usepackage{inconsolata}
\usepackage{times}
\usepackage{multirow}
\usepackage{graphicx}
\usepackage{booktabs} % added to preamble to define \addlinespace
\usepackage{url}
\title{Resources and Few-shot Learners for~In-context~Learning~in~Slavic~Languages}

\author{Michal Štefánik$^\diamondsuit$ \and
  Marek Kaldčík$^\diamondsuit$ \and
  Piotr Gramacki$^\clubsuit$ \and
  Petr Sojka$^\diamondsuit$ \\
  \\$^\diamondsuit$Faculty of Informatics,\\\vspace{10pt}Masaryk University, Czechia \\
  $^\clubsuit$Department of Artificial Intelligence,\\Wrocław University of Science and Technology, Poland
}
\begin{document}
\maketitle
\begin{abstract}

Despite the rapid recent progress in creating accurate and compact in-context learners, most recent work focuses on in-context learning (ICL) for tasks in English. 
However, the ability to interact with users of languages outside English presents a great potential for broadening the applicability of language technologies to non-English speakers.

In this work, we collect the infrastructure necessary for training and evaluation of ICL in a selection of Slavic languages\footnote{All our templates and models are available on \url{https://github.com/fewshot-goes-multilingual/slavic-incontext-learning}}: Czech, Polish, and Russian. We link a diverse set of datasets and cast these into a unified instructional format through a set of transformations and newly-crafted templates written purely in target languages.
Using the newly-curated dataset, we evaluate a set of the most recent in-context learners and compare their results to the supervised baselines. Finally, we train, evaluate and publish a set of in-context learning models that we train on the collected resources and compare their performance to previous work.

We find that ICL models tuned in English are also able to learn some tasks from non-English contexts, but multilingual instruction fine-tuning consistently improves the ICL ability. We also find that the massive multitask training can be outperformed by single-task training in the target language, uncovering the potential for specializing in-context learners to the language(s) of their application.
\end{abstract}

\section{Introduction}

The emergent ability of very large language models to understand unseen tasks from natural input text \citep{gpt3}, referred to as In-context Learning (ICL), recently motivated a large body of work focused specifically on creating more efficient models able to understand a new task from human instructions \cite{min-etal-2022-metaicl,sanh2022multitask,wei2022finetuned,chung2022_flan}. 
The ICL models presented in these works reduce the number of parameters compared to the first in-context learners by orders of magnitude. 
In exchange, they assume that the generalization to new tasks emerges from a vast mixture of diverse training tasks seen in the training process.
% While the ICL models presented in these works reduce the number of parameters of the first in-context learners by orders of magnitude, instead of the vast parametrization, these approaches assume that the generalization to new tasks emerges from a vast mixture of diverse training tasks seen in the training process.

The data volume and diversity requirements might also be the factor that substantially limits the application of current ICL models mainly to English.
Acquiring a large and diverse set of tasks is relatively easy for English, which is in the spotlight of the NLP community. 
Unfortunately, there are fewer datasets in other languages, and the collection of new ones is costly. 
%One is able to harvest a large and diverse set of tasks for English where the focus of the NLP community concentrates, but for other languages, the availability of datasets is sparser, and the collection of new ones is expensive. 
Previous work addresses this problem by automatic translation of some English datasets \cite{chandra_survey_nonenglish_QA}, or by a cross-lingual training \cite{mishra-etal-2022-cross} and evaluation~\cite{conneau2018xnli}.
However, such approaches do not resemble the use of instruction models by non-English speakers, expecting the models to interact \textit{solely} in their native language.

\begin{figure}[t]
\centerline{\includegraphics[width=0.975\linewidth]{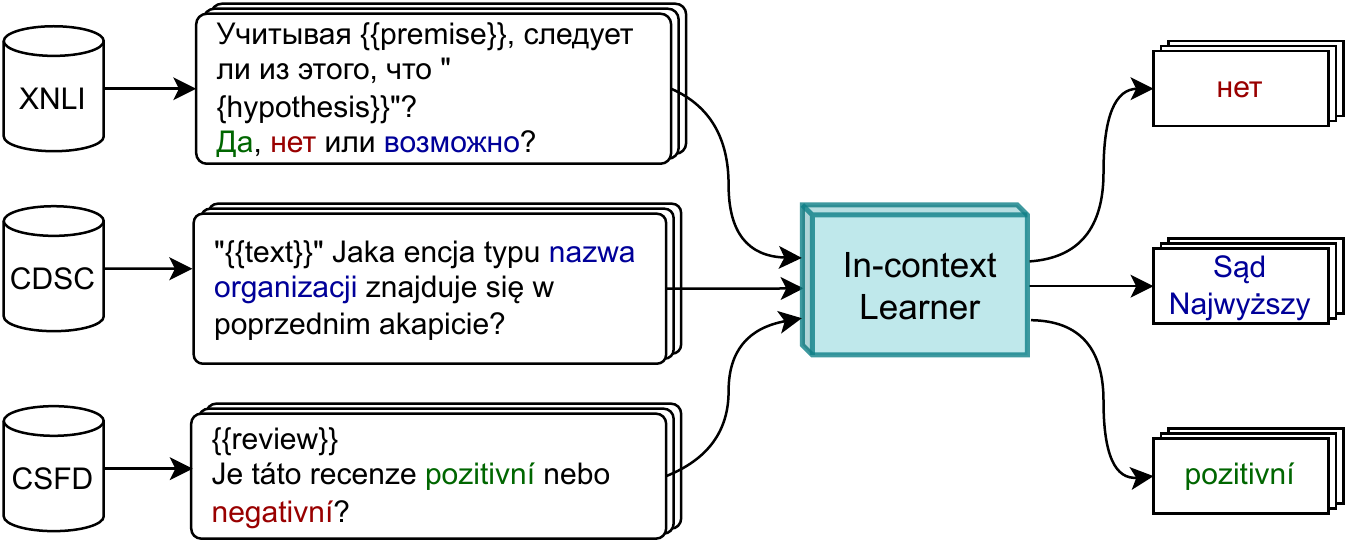}}%
% \vspace*{.2\baselineskip}
\caption{In this work, we transform Czech, Polish, and Russian datasets for diverse task types into a unified instructional format through a set of templates curated by the native speakers of target languages. The resulting collection enables an evaluation of existing in-context learners as well as the creation of new in-context learners interacting fully in the target language.}
% Image to edit: https://drive.google.com/file/d/1sxQmv0pa5tfOZb0eTwU8rj1wuyDhijRr/view?usp=sharing
\label{fig:abstract}
\end{figure}

This work evaluates the quality of in-context learning achievable in non-English languages to this date, specifically focusing on applicability in few-shot in-context learning for interaction in selected Slavic languages (Figure~\ref{fig:abstract}).
Further, we assess the possibilities of further improvement under the assumption of limited data availability in the target language.
We formulate these goals in two research questions:
\begin{description}
\item[\textbf{RQ1}:] \textit{How well can recent in-context few-shot learners \textbf{perform} in the interaction purely within our chosen, non-English languages?}

\item[\textbf{RQ2}:] \textit{Can the improvements of in-context learning in a large-resource language \textbf{transfer} to lower resource, target languages?}
\end{description}

Given very limited previous work in in-context learning in our target languages, within our work, we first (i)~survey and transfer a diverse set of datasets to instructional format through a set of transformations and newly-collected database of prompting templates with both the instructions and labels written in our target language(s).
Our collected tasks include datasets for Named Entity Recognition, Sentiment Classification, Natural language Inference, and Question Answering in our target languages.
After collecting the datasets of diverse tasks in the ICL-compatible format, we (ii)~survey and evaluate in-context few-shot learners that can be applied to our target languages.
Finally, we (iii)~explore the possibility of further improving the in-context learners specific for our target languages along two axes: (a)~by increasing models' exposure to target-language data and (b)~by improving ICL ability in high-resource language, evaluating the cross-lingual transfer of such improvements.

This paper is structured as follows.
Section~\ref{sec:background} overviews the standard settings of in-context few-shot learning and surveys the previous work in this direction.
Section~\ref{sec:dataset} describes the evaluation datasets that we use and covers datasets' selection and unification process and templates database collection. 
Section~\ref{sec:models} presents the settings used for training our in-context learners for Czech, Polish, and Russian.
Finally, Section~\ref{sec:results} presents the evaluation results, including existing and newly-trained in-context learners in the supervised baseline.

\section{Background}
\label{sec:background}

\paragraph{In-context learners}
In-context learning from both human prompt and a set of input-output examples is initially observed as an emergent ability of GPT-3 \cite{NEURIPS2020_brown_gpt3} trained on a vast collection of unlabelled texts for Causal language modeling (CLM) objective \cite{Radford2018gpt}.
Subsequent work reproduces ICL ability and open-sources the resulting models, such as BLOOM \cite{scao2022_bloom} or OPT \cite{zhang2022opt}.
However, in-context learners trained in a solely unsupervised fashion are impractically large and hence, expensive for conventional use; In unsupervised settings, the ICL ability seems to emerge only when using far over 10 billion parameters \cite{NEURIPS2020_brown_gpt3}, thus requiring an extensive infrastructure to perform a single inference.

Computational overhead is addressed by a series of smaller models trained \textit{specifically} for in-context learning. %The smaller in-context learners are trained using meta-learning approaches utilizing a large mixture of tasks in a consistent sequence-to-sequence format \cite{min-etal-2022-metaicl}, obtained using \textit{templates} \cite{bach2022promptsource}, that define the input prompts for each task in the collection.
The smaller in-context learners are trained with a large mixture of tasks converted to a consistent sequence-to-sequence format via human-written \textit{templates} \cite{bach2022promptsource} that define the input prompts for each task in the collection.
A popular use of this framework includes prefixing the input sequence with natural-language \textit{instructions}, such as the ones given to human annotators \cite{mishra-etal-2022-cross}. Large-scale instruction-based prompting in training over 1,600 tasks is also adopted in training \textsc{Tk-instruct} \cite{wang-etal-2022-super} that we assess in our evaluations.

Recently, more attention has been dedicated to a selection of in-context training \textit{tasks} under the assumption that some training tasks might be more beneficial for the emergence of in-context learning than others. In this direction, \textsc{FLAN-T5} of \citet{chung2022_flan} further extends a database of tasks with the ones requiring multi-step reasoning in a \textit{Chain-of-Thought} manner, where additionally to the correct prediction, the model is trained to predict a \textit{sequence of steps} mapping the input to an output.

\paragraph{In-context Few-shot learning} 
In-context learners are easily applicable in few-shot evaluation settings, where a small set of demonstrations for a given task exists.
Given a dataset $\mathcal{D}: \{(x_1\rightarrow Y_1),\ldots, (x_i\rightarrow Y_i)\} \in \mathcal{D}$ containing pairs of \textit{input} $x_j$ with associated \textit{label} $Y_j$, an \textit{in-context few-shot learner} $\Theta(x) \rightarrow y$ aims to predict a correct $y_{k+1} \equiv Y_{k+1}$ given \textit{input text} containing a sequence of $k$ input-output \textit{demonstrations}, and the predicted input $x_{k+1}$ \cite{stefanik2022_not_in_context,gao2022_pal}:
\begin{equation}
    \Theta([x_1\rightarrow Y_1,\ldots, x_k\rightarrow Y_k], x_{k+1}) \rightarrow y_{k+1}
    \label{eq:few_shot}
\end{equation}
Contrary to the standard supervised learning, in in-context learning, model $\Theta$ is \textit{not} updated. Thus, it can rely solely on its ability to understand the task from input text.

Similarly to humans, the specific wording of input, i.e., \textit{prompt} $x_j$, might play a large difference in the evaluation performance of the model. A~prompt formulation optimal for one model type is likely not optimal for another~\cite{lu-etal-2022-fantastically}. Therefore, in order to fairly compare different in-context learners, one should evaluate in-context learners on a larger set of diverse prompts~\cite{bach2022promptsource}.
With this motivation, we also collect multiple prompts for each task, with a focus on their mutual diversity.

\section{Datasets}
\label{sec:dataset}

The evaluation and training of new in-context learners for our target languages require (i)~a collection of datasets for a representative range of tasks, and (ii)~ the transformation of these datasets into a unified, self-containing sequence-to-sequence form of inputs and outputs.
Thus, one of our main contributions is the adaptation of the datasets for Czech, Polish, and Russian in a range of tasks: Named entity recognition, Sentiment classification, Natural language inference, and Question answering. The overview of the datasets for our target languages is shown in Table~\ref{tab:datasets}.

This section overviews the datasets in the target languages that we transformed, followed by a description of the process of constructing the templates for these datasets.

\begin{table}
\centerline{\scalebox{0.7}{
\begin{tabular}{@{}lllrc@{}}
\hline
\ & Name & \multicolumn{1}{l}{Task} & \multicolumn{1}{l}{Size} & \multicolumn{1}{@{}l@{}}{Templates} \\ \hline
\multirow{6}{*}{cs} & CNEC \cite{SevcikovaEtAl2007CNEC} & NER & 19k & 3 \\
 & CSFD (this work) & Clf. & 30k & 3 \\
 & FBCom \cite{Habernal.Brychcin.2013b} & Clf. & 7k & 3 \\
 & MALL \cite{Habernal.Brychcin.2013b} & Clf. & 30k & 3 \\
 & SQAD \cite{medved2022_sqad} & QA & 8k & 4 \\
 & CTKFacts \cite{ullrich2022_ctkfever} & NLI & 5k & 7 \\ \hline
\multirow{4}{*}{pl} & PoliticAds \cite{augustyniak-etal-2020-political} & NER & 1k & 4 \\
 & KPWR \cite{broda-etal-2012-kpwr} & NER & 9k & 4 \\
 & Polemo \cite{kocon-etal-2019-multi} & Clf. & 8k & 4 \\
 & CDSC \cite{wroblewska-krasnowska-kieras-2017-polish} & NLI & 10k & 4 \\ \hline
\multirow{4}{*}{ru} & Polyglot \cite{polyglotner} & NER & 136k & 3 \\
 & CEDR \cite{sboev2021data} & Clf. & 9k & 3 \\
 & SberQuAD \cite{efimov2019_sberquad} & QA & 74k & 4 \\
 & XNLI \cite{conneau2018xnli} & NLI & 399k & 7 \\ 
 \hline
\end{tabular}
}}
\caption{Overview of datasets that we transform to a sequence-to-sequence format through manually-crafted templates in target languages.}
\label{tab:datasets}
\end{table}

\subsection{Data Collections in Target Languages}

Contrary to English, labelled resources in our target languages for some tasks are relatively sparse, which conditions us to undertake some compromises in the diversity of the resources that we proceed with.
The following text also covers the transformation that we had to perform with these datasets to cast them into a unified sequence-to-sequence format.

\subsubsection{Czech Datasets} Contrary to Polish with a larger base of speakers, Czech datasets include all tasks that we aim to collect, including NER, Classification, QA, and NLI. 

\textbf{CNEC} \cite{SevcikovaEtAl2007CNEC} dataset for \textbf{NER} presents entities in the context of radio transcripts and news articles, featuring a relatively large collection of more than 10,000 original texts. We transform this dataset into sequence-to-sequence form by querying a specific type of entity, where we only use samples containing at most one occurrence of the requested entity to avoid ambiguity.

We note that all \textbf{classification} datasets that we find for evaluation are focused on a specific case of sentiment classification.
Nevertheless, the volume, quality, and variance of sentiment classification datasets are relatively high;
(i)~\textbf{CSFD} presents a set of 30,000 public reviews from the movie critiques with diverse vocabulary and the challenging end task of predicting the corresponding star rating (0--5).
The dataset is balanced, with each rating having a similar number of occurrences.
To evaluate the models in a natural language, instead of predicting a specific numeric rating for each review, we transform the dataset labels to \textit{positive/negative} classification, omitting samples with rating=3.
(ii)~\textbf{MALL} \cite{Habernal.Brychcin.2013b} dataset is a semantically less complex collection of product reviews of online store products, and 
(iii)~\textbf{FBCom} \cite{Habernal.Brychcin.2013b} features a collection of scraped but verified Facebook comments presenting a sample of informal language.
The latter two datasets come with three-class targets (positive/neutral/negative).

The only available Czech \textbf{QA} dataset, \textbf{SQAD} \cite{medved2022_sqad}, also builds a dataset on Wikipedia, containing the original articles in a full length, associated with manually-crafted questions and associated answer texts. To avoid the overhead of models' inference with full Wikipedia articles in a few-shot format, we synthesize the contexts containing answers by sequencing paragraphs containing the first answer occurrence.
Thus, our curated context paragraphs resemble the format of the commonly-known English SQuAD dataset \cite{rajpurkar-etal-2016-squad}.
We note that the original version of the dataset contains a strong statistical bias, with around half of the questions having the answer at the beginning of the article. 
To avoid exploiting this bias in evaluation, we randomly removed 90\% of the questions whose answer starts in the first 50~characters.

Finally, \textbf{CTKFacts} \cite{ullrich2022_ctkfever} introduces a collection of \textbf{NLI} examples containing premises extracted from Wikipedia, with manually-crafted hypotheses to assess given the premises, in standard NLI settings. 

\subsubsection{Polish}

The Polish datasets for our desired tasks are smaller than Czech, and contrary to Czech, to the date of writing, we find no publicly-available Polish QA dataset. 
However, we find two Polish \textbf{NER} datasets. \textbf{PoliticAds} \cite{augustyniak-etal-2020-political} presents input texts in a relatively unconventional domain of political advertising.
A lot of entities are largely context-dependent, thus presenting adaptation challenges for general-domain models.
Therefore, we complement this quite small and specific dataset with the \textbf{KPWR} \cite{broda-etal-2012-kpwr} dataset. 
However, original KPWR has a very fine granularity of entities; thus, we transform the target entities to a second-level type (i.e. mapping entity \textit{name-location-city} simply to \textit{location}).
After disambiguation analogical to CNEC, we obtain a sequence-to-sequence dataset with 9,000 inputs.

Consistently to Czech, we enrich the set with \textbf{Polemo} dataset \cite{kocon-etal-2019-multi} for sentiment \textbf{classification}, which contains a human-annotated set of consumer reviews from the domains of \textit{medicine}, \textit{hotels}, \textit{products}, and \textit{university}. 
Finally, we find \textbf{CDSC} dataset for \textbf{NLI} \cite{wroblewska-krasnowska-kieras-2017-polish}, featuring a collection of premise-hypothesis pairs from a wide range of 46~thematic groups.

\subsubsection{Russian}

Being the language with a much larger speaker base, Russian is also the richest in resources. Thus, we pick the datasets for our tasks of interest that we assess as having the highest quality. \textbf{Polyglot} \cite{polyglotner} is a large \textbf{NER} dataset curated from references to Wikipedia sites.
We transform the datasets to per-entity-type prompt format, creating multiple prompts from each sample, resulting in more than 100\,k input-output entity pairs. Consistently with other languages, we further include in the collection a \textbf{CEDR} dataset for sentiment \textbf{classification} originating in social media \cite{sboev2021data}.
While its domain is not representative of many use cases, we assess the quality of annotations as superior to its alternatives and the number of labels (5) as practical for few-shot evaluation with reasonably long contexts.

\textbf{SberQuAD} \cite{efimov2019_sberquad} is an extractive \textbf{QA} dataset comparable with English SQuAD in both the size and domain; Its 74,000 question-context-answer tuples are manually collected with the contexts originating in Wikipedia.
Contrary to SQuAD, a small portion of questions has several different answers in the context, making the correct prediction ambiguous in some cases; We omit these cases in evaluations.
Finally, we choose an \textbf{XNLI} dataset \cite{conneau2018xnli} for evaluating \textbf{NLI} in Russian for its heterogeneity and size.
However, other quality alternatives exist (see, e.g. \citet{shavrina2020russiansuperglue}), and our templates can be used with any other Russian NLI dataset as well.

\subsection{Templates}
For each of the referenced datasets, we write a new template mapping the samples of the dataset into a sequence-to-sequence format.
To reinforce templates' heterogeneity, we start by reviewing existing templates of the analogical tasks in English, collected within BigScience's P3 project \cite{sanh2022multitask}.
From existing templates, we pick a set of mutually most-distinct templates for each task and proceed to the writing phase.
The resulting number of templates for each dataset was chosen subjectively to maintain a high level of heterogeneity among the templates of each dataset.

Inspired by the existing templates, we ask our target-language volunteer native speakers to write the templates in a form that they find ``the most natural to ask for the solution for a given task from a human with a native understanding of their target language''.
We make sure that all the templates contain the exact-matching form of the expected response (i.e., \textit{label}) so that the domain of possible answers is clearly enclosed by the prompt.
The examples of some curated templates can be found in Table~\ref{tab:example_templates}.
A~full list of the collected templates can be found in Appendix~\ref{appx:templates}.

\def\RUS#1{\fontfamily{qtm}\fontencoding{T2A}\selectfont#1}
\begin{table*}
\renewcommand{\baselinestretch}{0.94}
 \centerline{\scalebox{.76}{%
 \begin{tabular}{@{}l@{\,}l@{\ \ }p{1.2\textwidth}@{}}
 \toprule
 Lang & Task & Template\\
\midrule
 cs       & NER  & \{\{text\}\} Jaká entita typu \{\{label\_type\}\} se nachází v předchozím odstavci? \\                                                                                 
 cs       & Clf. & \{\{comment\}\} Je tato recenze \{\{"pozitivní, neutrální nebo negativní"\}\}? \\                                                                                      
 cs       & QA   & \{\{context\}\}  Q: \{\{question\}\}  S odkazem na sekci výše je správná odpověď na danou otázku                                                                      \\
 cs       & NLI  & Za předpokladu, že \{\{evidence\}\} vyplývá, že \{\{claim\}\}? Ano, ne, nebo možná?                                                                                                                        \\
 %\midrule
 \addlinespace
 pl       & clf. & "\{\{text\}\}" Ten tekst jest pozytywny, negatywny, neutralny czy dwuznaczny? \\
 pl       & NLI  & Oceń czy poniższe zdania są zgodne ze sobą - tak, nie czy nie wiadomo?   Zdanie A: \{\{premise\}\}  Zdanie B: \{\{hypothesis\}\}   Zgodność: \\
 pl       & NER  & Jaka encja typu \{\{label\_type\_selected\}\} znajduje się w następującym tekście? "\{\{text\}\}" \\
 \addlinespace
 ru       & NER  & \{\{text\}\} \RUS{Какой объект типа} \{\{label\_type\}\} \RUS{находится в предыдущем абзаце}? \\
 ru       & NLI  & \RUS{Примите за истину следующее:} \{\{premise\}\} \RUS{Тогда следующее утверждение:} "\{\{hypothesis\}\}" \RUS{есть "правда", "ложь" или "неубедительно"}? \\
 ru       & QA   & \RUS{Посмотрите на абзац ниже и ответьте на следующий вопрос:  Абзац:} \{\{context\}\}  \RUS{Вопрос:} \{\{question\}\} \\
 ru       & Clf. & \{\{text\}\} \RUS{Каково настроение этого обзора? радость, печаль, удивление, страх или гнев?} \\
 \bottomrule
 \end{tabular}}}
 \caption{Examples of instruction templates for each of the language + task pair that we collect in this work. A full list of templates collected in this work by our native speakers can be found in Appendix~\ref{appx:templates} Table~\ref{tab:all_templates}.}
\label{tab:example_templates}
\end{table*}

We do not identify any instructional templates for the Named Entity Recognition task in the previous work.
This is likely due to the complexity of fair evaluation of prediction containing a \textit{sequence} of prediction, necessary for collecting \textit{all} predictions for the prompted entity type; an evaluation of sequences is difficult by using the commonly-used generative measures.
After consideration, we decided to reformulate the NER tasks in the form of information extraction, where we filter out samples where prompted entity type occurs more than once.
This makes the task easier, but on the other hand, the evaluation is not biased by the models' ability to order predictions correctly.
Based on that, we assume that such evaluation corresponds better to in-context learners' ability to identify entities.
%This makes the in-context learning task easier, but the evaluation is not biased by the models' ability to correctly order prediction, making us assume that such evaluation corresponds better to in-context learners' ability to identify entities.

\section{Experiments}
\label{sec:models}

Making in-context learning in our target languages finally possible through the transformations described in the previous section, our first objective is to assess the current state-of-the-art of the recent in-context few-shot learners when used in the interaction \textit{exclusively} in the target language (\textbf{RQ1}). We follow by outlining the perspectives in further enhancing the quality of target-language in-context few-shot learners by assessing the potential of cross-lingual transfer (\textbf{RQ2}).

\subsection{In-context Few-shot Learning Evaluation}
\label{sec:icl_evaluation}

The overview of previous work on in-context learning covered in Section~\ref{sec:background} shows a shifting interest from the over-parametrization to the scaling of diverse training tasks \cite{wang-etal-2022-super} and more explicit reasoning schemes, such as a Chain-of-Thought \cite{chung2022_flan}, where in addition to the final result, the model learns to predict the reasoning path that has led to the prediction.
Our evaluation aims to assess how these aspects impact the quality of in-context few-shot learning in our target languages.

\paragraph{Multilingual fine-tuning} To this date, we identify only one in-context learners' family that claims to support all our target languages: \textsc{mTk-instruct} \cite{wang-etal-2022-super}. While its English counterpart (\textsc{Tk-instruct}) fine-tunes \textsc{T5} models \cite{t5} on 1,616 tasks with English prompts, inputs, and targets, \textsc{mTk-instruct} is additionally fine-tuned on 576~tasks with inputs in 55~diverse languages, including Czech, Polish and Russian. Still, the instructional templates for these languages were written in English due to easier quality assurance. Thus, it remains an open question whether such-acquired in-context learning skills transfer to an interaction \textit{solely} in the target language.

Hence, we assess the benefit of multilingual training by measuring and comparing the performance of English-only \textsc{Tk-instruct} and multilingual \textsc{Tk-instruct} of the same size (3\,B parameters).

\paragraph{Fine-tuning strategy}
We evaluate the impact of a set of objectives of \textsc{FLAN-T5} \cite{chung2022_flan} complementary to a sole scaling of tasks of \textsc{Tk-instruct}. Notably, these include (i)~additional fine-tuning for a zero-shot setting, i.e. without presenting the model with demonstrations, (ii)~fine-tuning for generating Chain-of-Thought, i.e. a sequence of steps leading the model to the answer, that is purposed to enhance the model's reasoning ability.

The evaluations of the impact of a fine-tuning strategy are also complemented by the assessment of our newly-trained in-context learners, trained on a single task type (QA), including the data in a target language; We detail our approach to train these models in Section~\ref{sec:cross-lingual_transfer}.

\paragraph{Model size}
Finally, we evaluate both \textsc{Tk-instruct} and \textsc{FLAN-T5} in two different sizes: in a 700-million and in a four-times bigger, 3-billion-parameters variant. While it is perhaps not a surprising finding that the larger model would also perform better in the unseen language, the experiments in this axis assess the scale of improvement that can be expected by increasing computational costs for larger models, as compared to other adjustments.

\subsection{Cross-lingual Transfer}
\label{sec:cross-lingual_transfer}

In addition to the evaluation of existing in-context learners, we are interested in assessing how much the ICL in lower-resource languages can benefit from the improvements in a large-resource language (\textbf{RQ2}). This is particularly relevant given the fast pace of progress in general in-context learning focused primarily on English, naturally raising a question on how applicable these results are in languages for which data resources are sparser.

However, having no control over the specific data and training configuration of the existing models, we assess the scale of cross-lingual transfer by fine-tuning our own in-context learners that differ in the configuration in a large-resource language (English) while fixing the configuration in the target language. By also considering the choices of the previous work \cite{sanh2022multitask}, we pick the Question Answering as the one that we assume is crucial for obtaining in-context learning ability while also being available in our target languages.

Therefore, in our experiments, we \textit{permute} only the English QA dataset and mix it in training with the QA dataset of the target language. We train in-context learners with three different configurations; (i)~using \textit{no} English QA dataset, (ii)~using the standard SQuAD \cite{rajpurkar-etal-2016-squad} containing more than 90,000 question-context-answer tuples, and (iii)~using a lesser-known AdversarialQA (AQA) dataset \cite{bartolo-etal-2021-improving} containing 30,000 more complex questions that \textit{exploit} the flaws of QA models trained on SQuAD, making its samples complementary to SQuAD. Finally, we measure the impact of this change in Czech and Russian, for which the target-language QA datasets are available.

All our newly-trained in-context learners (further referred to as m$\textsc{Tk-QA}_{SQuAD}$ and m$\textsc{Tk-QA}_{AQA}$) are based on m\textsc{T5} model \cite{xue-etal-2021-mt5} of 1.3-billion parameter size. We make our newly-trained in-context learners for both Czech\footnote{\url{https://huggingface.co/fewshot-goes-multilingual/mTk-SQuAD_en-SQAD_cs-1B}} and Russian\footnote{\url{https://huggingface.co/fewshot-goes-multilingual/mTk-AdversarialQA_en-SberQuAD_ru-1B}} publicly available for any use.

\def\tabscale{0.75}
\def\mytabcolsep{3dd}
\begin{table*}[t]
\tabcolsep\mytabcolsep
\centerline{\scalebox{\tabscale}{%
\begin{tabular}{lllllll}
\toprule
\multicolumn{1}{r}{Dataset + task} & \multicolumn{1}{c}{CNEC} & \multicolumn{1}{c}{CSFD} & \multicolumn{1}{c}{FBCom} & \multicolumn{1}{c}{MALL} & \multicolumn{1}{c}{SQAD} & \multicolumn{1}{c}{CTKFacts} \\
\multicolumn{1}{l}{Model} & \multicolumn{1}{c}{NER} & \multicolumn{1}{c}{Clf.} & \multicolumn{1}{c}{Clf.} & \multicolumn{1}{c}{Clf.} & \multicolumn{1}{c}{QA} & \multicolumn{1}{c}{NLI} \\
\midrule
Supervised (mT5-1B) & 67.9±\ \,9.1 & 82.4±4.5 & 49.3±10.3 & 42.8±10.8 & 88.3±5.3 & 56.1±10.9 \\
\addlinespace

\textsc{Tk}-Instruct (700M) & 15.3±\ \,6.7 & 14.1±7.1 & 25.2±\ \,7.2 & 25.5±\ \,8.4 & \ \,5.6±4.8 & \textbf{54.7}±8.2 \\
\textsc{Tk}-Instruct (3B) & 32.8±\ \,9.1 & 20.9±8.1 & 23.0±\ \,7.4 & 25.1±\ \,6.9 & 34.0±9.0 & 47.8±9.8 \\
T5-FLAN (700M) & 41.1±10.0 & \ \,0.0±\ \,0.0 & \ \,0.0±\ \,0.0 & \ \,0.0±\ \,0.0 & 46.5±8.4 & 30.3±9.3 \\
T5-FLAN (3B) & 49.6±10.4 & \ \,0.0±\ \,0.0 & \ \,0.0±\ \,0.0 & \ \,0.1±\ \,0.1 & 51.6±9.1 & 34.7±10.7 \\
\addlinespace

m\textsc{Tk}-Instruct (3B) & 62.5±\ \,8.9 & \textbf{90.2}±4.2 & 10.8±6.2 & \ \,9.9±7.0 & 67.9±8.6 & 44.0±10.1 \\
\addlinespace

m$\textsc{Tk-QA}_\textit{none}$(1B) & 72.0±\ \,9.0 & 45.9±9.1 & 29.2±8.2 & 32.1±8.9 & 85.0±7.0* & 35.4±10.5 \\
m$\textsc{Tk-QA}_\textsc{SQuAD}$(1B) & 72.3±\ \,9.1 & 72.9±6.6 & \textbf{32.1}±9.0 & \textbf{34.7}±9.2 & \textbf{87.8}±5.3* & 46.9±10.1 \\
m$\textsc{Tk-QA}_\textsc{AQA}$(1B) & \textbf{77.0}±\ \,7.8 & 59.8±8.8 & 27.6±8.6 & 29.8±9.9 & 87.1±6.6* & 42.7±10.7 \\
\bottomrule
\end{tabular}}}

\caption{\textbf{In-context learners' performance in Czech:} ROUGE-L scores of selected in-context learners in Czech interaction using the listed datasets, for the best-performing template of each model. In-context learners were shown \textbf{three} demonstrations of each task. Included confidence intervals ($\alpha=0.05$) are computed using bootstrapped evaluation (sample groups $n=100$, repeats $r=200$). Results marked with * denote cases where the held-out set of the listed dataset was used in training.}
\label{tab:results_czech}
\end{table*}

\begin{table}[t]
\tabcolsep\mytabcolsep
\centerline{\scalebox{\tabscale}{%
\begin{tabular}{@{}lllll@{}}
\toprule
\multicolumn{1}{r}{Dataset + task} & \multicolumn{1}{c}{PoliticAds} & \multicolumn{1}{c}{KPWR} & \multicolumn{1}{c}{Polemo} & \multicolumn{1}{c}{CDSC} \\
\multicolumn{1}{@{}l}{Model} & \multicolumn{1}{c}{NER} & \multicolumn{1}{c}{NER} & \multicolumn{1}{c}{Clf.} & \multicolumn{1}{c}{NLI} \\ 
\midrule

Supervised (mT5-1B) & \ \,5.9±5.1 & 63.3±10.3 & 51.9±9.9 & 75.5±8.5 \\

% \hline
\addlinespace

\textsc{Tk}-Instruct (700M) & \ \,5.6±4.3 & \ \,8.6±5.4 & 28.3±8.6 & 52.3±8.2 \\
\textsc{Tk}-Instruct (3B) & 17.6±8.1 & 54.6±11.2 & 19.5±8.4 & 67.8±8.8 \\
T5-FLAN (700M) & \ \,6.8±5.5 & 33.8±9.8 & 24.3±8.6 & 10.0±6.4 \\
T5-FLAN (3B) & 18.4±7.3 & 60.5±7.8 & \textbf{43.0}±9.0 & \textbf{71.5}±9.0 \\
\addlinespace

m\textsc{Tk}-Instruct (3B) & \textbf{32.1}±9.6 & \textbf{67.6}±8.4 & 25.4±8.6 & 70.6±8.2 \\
\bottomrule
\end{tabular}}}
\caption{\textbf{In-context learners' performance in Polish:} ROUGE-L scores of selected in-context learners in Polish interaction using the listed datasets. Configuration of evaluation is identical to Table~\ref{tab:results_czech}.}
\label{tab:results_polish}
\end{table}

\begin{table}[t]
\tabcolsep\mytabcolsep
\centerline{\scalebox{\tabscale}{%
\begin{tabular}{@{}llll@{}l@{}}
\toprule
\multicolumn{1}{r}{Dataset + task} & \multicolumn{1}{c}{Polyglot} & \multicolumn{1}{c}{CEDR} & \multicolumn{1}{c}{SberQAD} & \multicolumn{1}{c}{XNLI} \\
\multicolumn{1}{@{}l}{Model} & \multicolumn{1}{c}{NER} & \multicolumn{1}{c}{Clf.} & \multicolumn{1}{c}{QA} & \multicolumn{1}{c}{NLI} \\
\midrule
Supervised (mT5-1B)& 54.3±10.8 & 48.6±9.6 & 86.4±6.5 & 51.5±11.5 \\
\addlinespace

\textsc{Tk}-Instruct (700M) & \ \,0.1±0.5 & 12.2±6.8 & \ \,0.6±1.1 & 12.9±\ \,6.9 \\
\textsc{Tk}-Instruct (3B) & \ \,3.6±3.9 & 17.7±8.3 & \ \,8.1±4.1 & 22.2±\ \,8.2 \\
T5-FLAN (700M) & \ \,1.0±1.6 & 15.1±6.1 & 11.4±4.8 & 13.8±\ \,6.2 \\
T5-FLAN (3B) & \ \,2.0±2.5 & 24.4±7.4 & 19.6±5.6 & 26.0±\ \,9.0 \\
\addlinespace

m\textsc{Tk}-Instruct (3B) & 57.6±11.2 & \textbf{33.0}±9.9 & 73.7±6.7 & \textbf{35.3}±10.3 \\
\addlinespace

m$\textsc{Tk-QA}_\textit{none}$(1B) & 53.3±8.4 & 17.9±8.1 & \textbf{89.1}±5.2* & 19.6±\ \,7.5 \\
m$\textsc{Tk-QA}_\textsc{SQuAD}$(1B) & 50.3±9.3 & \ \,7.5±4.5 & 84.6±6.0* & 23.8±\ \,8.8 \\
m$\textsc{Tk-QA}_\textsc{AQA}$(1B) & \textbf{66.3}±10.9 & 27.0±9.9 & 86.0±5.6* & 32.3±\ \,8.3 \\
\bottomrule
\end{tabular}}}
\caption{\textbf{In-context learners' performance in Russian:} ROUGE-L scores of selected in-context learners in Russian interaction using the listed datasets. Configuration of evaluation is identical to Table~\ref{tab:results_czech}.}
\label{tab:results_russian}
\end{table}

\section{Results}
\label{sec:results}

Consistently with the previous work \cite{sanh2022multitask,wang-etal-2022-super}, we jointly report the ROUGE-L score \cite{lin-2004-rouge} over all the evaluation datasets (which we transform and create templates for (§\ref{sec:dataset})) and all the evaluated in-context learners (§\ref{sec:icl_evaluation}), including the newly-trained ones introduced in this work (§\ref{sec:cross-lingual_transfer}). To ease the readability, we split the reports by language, to the results on Czech datasets in Table~\ref{tab:results_czech}, Russian datasets in Table~\ref{tab:results_russian}, and Polish datasets in Table~\ref{tab:results_polish}.

As a reference of the resulting ICL performance, for each dataset, we also train a \textbf{baseline model} that is also based on m\textsc{T5} model \cite{xue-etal-2021-mt5}, fine-tuned on the training split of the dataset transformed to a sequence-to-sequence format through a mixture of \textit{all} the templates that we curated. Details on the training and evaluation configuration that we use can be found in Appendix~\ref{appx:details}.

\paragraph{Multilingual training helps in most cases}
A comparison of m\textsc{Tk-Instruct} to \textsc{Tk-Instruct} of the same size through all languages (Tables~\ref{tab:results_czech}, \ref{tab:results_russian}, \ref{tab:results_polish}) evaluates the significance of including the training data from the target language(s). Note that mT5, a base model for mTk-instruct, was pre-trained on mC4 balanced over languages, but mTk-instruct was finetuned on only 15~Polish, 5~Russian, and 2~Czech datasets making it about 1\% of all data. Additionally, the training prompts for these datasets were English.

Still, we see that m\textsc{Tk-Instruct} is better than its English-finetuned counterpart in all evaluation datasets, except two Czech sentiment classification tasks. However, in some cases, the differences are relatively small; For instance, in the case of Polish CDSC, where English Tk-Instruct ends only 2.8~points behind the multilingual counterpart. The error analysis of mTk-Instruct on two flawing classification tasks (FBCom and MALL) has shown that despite purely Czech prompts, the model generates English responses. This could be explained by a semantic similarity of our tasks to some of the model's fine-tuning datasets, but in our evaluation, we consider the divergence from the prompted language of interaction a valid failure.

\paragraph{Inconsistent benefits of CoT training}
Comparing the performance of T5-FLAN models with Tk-instruct models of the corresponding size, we find that T5-FLAN is superior in 17 out of 28~cases. However, the differences are often relatively small, and the performance of both in-context learners in these cases remains below the usable level nevertheless. Therefore, while it seems that fine-tuning to a Chain-of-Thought reasoning allows the modeling of features that are applicable also in some multilingual settings, these do not generalize over all in-context learning scenarios. Notably, T5-FLAN perhaps surprisingly fails on classification in Czech, where it shows an inability to understand the task even from the given demonstrations. On the other hand, we note that in two of four evaluation cases in Polish, the larger T5-FLAN performs superiorly to even multilingual mTk-Instruct of the same size.

\paragraph{Model size matters}
The comparisons of T5-FLAN and Tk-Instruct in their two size variants show the superiority of the larger model with the exceptions in 3 out of 28~cases, suggesting that model size can be an even more important condition of accurate in-context learning ability than utilization of target-language data in training.

It is also worth noticing that the difference in performance between two \textit{sizes} of T5-FLAN are often very large; For instance, note the difference between Polish CDSC or Russian NLI. This suggests that the different sizes of T5-FLAN might, in fact, be very distinct in their representations.

\paragraph{Cross-lingual transfer}
A comparison of m\textsc{Tk-QA} models that we train with and without the high-resource QA dataset (§\ref{sec:cross-lingual_transfer}) outlines the potential for improvement of ICL in lower-resource languages with adjustments in the high-resource language. We see that including a complementary QA dataset in other-than-evaluated language can help in in-context learning of \textit{all} new tasks, with improvements over 60\% in Czech CSFD, or Russian XNLI.

Additionally, using a higher-quality AdversarialQA can also significantly, though not consistently, improve ICL ability for some tasks.
For instance, note the difference of 12.9 points in sentiment classification of the Czech CSFD dataset or of 16 in Russian NER. This relatively large sensitivity to the data configuration in a high-resource language, from which we aim to transfer the ICL ability, suggests that recent and future improvements in models' ICL measured in English might also be directly applicable to other languages.

\paragraph{In-context learners trained on a single task are comparable to multi-task learners}
While outperforming the in-context learners trained on a much larger scale of tasks was not our initial objective, we note that at least one of our in-context learners trained using a single (QA) task \textit{out-performs} mTk-Instruct in 6 out of 10~Czech and Russian evaluations. In \textit{all} other cases, a QA model performs within the confidence interval of mTk-Instruct. Additionally, in 4 out of 10 cases, at least one of our QA models performs comparably or better than the supervised baseline. Hence, rather than a weak performance of mTk-Instruct, this result underlines the efficiency of Question answering as a proxy task for generalizing to the unseen tasks. We also find this result encouraging for creating in-context learners specialized to other target languages, with a perspective to outperform generic state-of-the-art learners in a similar methodology.

\section{Conclusion}
\label{sec:conclusion}

This paper documents our work in creating the evaluation benchmark for in-context learning for Czech, Polish, and Russian. We transform selected datasets into a compatible format, and with the aid of volunteer native speakers, we create templates for these datasets exclusively in the evaluated language. However, our templates can be applied to any other dataset of the supported types (NER, Classification, QA, and NLI).

In the interaction that is purely in the language(s) of our interest, we evaluate a set of recent in-context learners that we consider state-of-the-art in this area. We find that even in-context learners trained dominantly on English data might perform considerably well and even outperform a fully supervised baseline in some cases. However, on average, massive multilingual pre-training and instruction-based fine-tuning still largely improve the ICL ability.

Finally, we train a set of in-context learners specifically for our target languages by mixing the large QA datasets in English with smaller QA datasets in our target languages; In both Czech and Russian, such-created learners perform better or comparably to mTk-Instruct trained on a vastly larger collection of over 2,000 tasks from 55~languages. We believe that this finding will motivate future work in creating specialized but more accurate in-context learners also for other languages outside English.

We publicly release all data transformations, templates, and the newly-created in-context learners for any use.

\section*{Limitations}
\paragraph{Templates}
While the templates that we curate with the help of native speakers were picked to maximize their mutual diversity, we acknowledge that the volumes of templates that we create for some datasets do not cover the full variance of possible prompts of our tasks. Therefore, our templates might not be optimal for our evaluated in-context learners.

\paragraph{Models}
In-context learners fine-tuned specifically for in-context instruction learning, including our introduced ones, are orders of magnitude smaller than the original language models acquired from sole pre-training like 175-billion-parameter GPT-3 \cite{NEURIPS2020_brown_gpt3}, but still remain compute-demanding for widespread deployment; We notice the inference time of a single sample for our 1B models to range between 3 and 10~seconds on a four-core CPU typical for middle-level personal computers to this date.

Analogically, also the application of our methodology (§\ref{sec:cross-lingual_transfer}) to other languages with similar size of the base model (1.3\,B) constrains the users to use dedicated GPU hardware with a minimum of 30\,GB memory. We train our assessed in-context learners using Nvidia A100 GPUs with 80\,GB VRAM, where the convergence of a single mT5-based model takes approximately 40~hours of computing.

\subsection*{Acknowledgements}

We acknowledge the Centre for Biomedical Image Analysis at Masaryk
University supported by MEYS CR (LM2018129 and
CZ.02.1.01/0.0/0.0/18\_046/0016045 Czech-BioImaging) for their support
in obtaining the results presented in this paper.

We are also grateful for the support of Kirill Semenov from Charles University, Prague, for willingly creating and proofreading newly-collected Russian templates as a native speaker.

% Biblio: 2 pages max
%TODO Petr (fix the references with sooo many authors to fit 2 page constraint), probably by https://tex.stackexchange.com/questions/433025/limit-author-list-in-creference-using-natbib
% shortening on .bib side as last resort: https://gist.github.com/zimmerst/9cb2ccad69b5f55a0a222c01b1d8e183

\bibliography{stefanik}

\begin{thebibliography}{36}
\expandafter\ifx\csname natexlab\endcsname\relax\def\natexlab#1{#1}\fi

\bibitem[{Al-Rfou et~al.(2015)Al-Rfou, Kulkarni, Perozzi, and
  Skiena}]{polyglotner}
Rami Al-Rfou, Vivek Kulkarni, Bryan Perozzi, and Steven Skiena. 2015.
\newblock \href {https://arxiv.org/pdf/1410.3791.pdf} {{Polyglot-NER}: Massive
  multilingual named entity recognition}.
\newblock \emph{{Proc. of the 2015 {SIAM} International Conference on Data
  Mining, Vancouver, British Columbia, Canada}}.

\bibitem[{Augustyniak et~al.(2020)Augustyniak, Rajda, Kajdanowicz, and
  Bernaczyk}]{augustyniak-etal-2020-political}
Lukasz Augustyniak, Krzysztof Rajda, Tomasz Kajdanowicz, and Micha{\l}
  Bernaczyk. 2020.
\newblock \href {https://www.aclweb.org/anthology/2020.winlp-1.28} {{Political
  Advertising Dataset: the use case of the Polish 2020 Presidential
  Elections}}.
\newblock In \emph{Proceedings of the The Fourth Widening Natural Language
  Processing Workshop}, pages 110--114, Seattle, USA. ACL.

\bibitem[{Bach et~al.(2022)Bach, Sanh, Yong, Webson, Raffel, Nayak, Sharma,
  Kim, Bari, Fevry, Alyafeai, Dey, Santilli, Sun, Ben-David, Xu, Chhablani,
  Wang, Fries, Al-shaibani, Sharma, Thakker, Almubarak, Tang, Tang, Jiang, and
  Rush}]{bach2022promptsource}
Stephen~H. Bach, Victor Sanh, Zheng-Xin Yong, Albert Webson, Colin Raffel,
  Nihal~V. Nayak, Abheesht Sharma, Taewoon Kim, M~Saiful Bari, Thibault Fevry,
  Zaid Alyafeai, Manan Dey, Andrea Santilli, Zhiqing Sun, Srulik Ben-David,
  Canwen Xu, Gunjan Chhablani, Han Wang, Jason~Alan Fries, Maged~S.
  Al-shaibani, Shanya Sharma, Urmish Thakker, Khalid Almubarak, Xiangru Tang,
  Xiangru Tang, Mike Tian-Jian Jiang, and Alexander~M. Rush. 2022.
\newblock \href {http://arxiv.org/abs/2202.01279} {Promptsource: An integrated
  development environment and repository for natural language prompts}.

\bibitem[{Bartolo et~al.(2021)Bartolo, Thrush, Jia, Riedel, Stenetorp, and
  Kiela}]{bartolo-etal-2021-improving}
Max Bartolo, Tristan Thrush, Robin Jia, Sebastian Riedel, Pontus Stenetorp, and
  Douwe Kiela. 2021.
\newblock \href {https://doi.org/10.18653/v1/2021.emnlp-main.696} {Improving
  question answering model robustness with synthetic adversarial data
  generation}.
\newblock In \emph{Proceedings of the 2021 Conference on Empirical Methods in
  Natural Language Processing}, pages 8830--8848, Online and Punta Cana,
  Dominican Republic. Association for Computational Linguistics.

\bibitem[{Broda et~al.(2012)Broda, Marci{\'n}czuk, Maziarz, Radziszewski, and
  Wardy{\'n}ski}]{broda-etal-2012-kpwr}
Bartosz Broda, Micha{\l} Marci{\'n}czuk, Marek Maziarz, Adam Radziszewski, and
  Adam Wardy{\'n}ski. 2012.
\newblock \href
  {http://www.lrec-conf.org/proceedings/lrec2012/pdf/965_Paper.pdf} {{KPW}r:
  Towards a free corpus of {P}olish}.
\newblock In \emph{Proc. of the Eighth International Conference on Language
  Resources and Evaluation ({LREC}'12)}, pages 3218--3222, Istanbul, Turkey.
  European Language Resources Association (ELRA).

\bibitem[{Brown et~al.(2020{\natexlab{a}})Brown, Mann, Ryder, Subbiah, Kaplan,
  Dhariwal, Neelakantan, Shyam, Sastry, Askell, Agarwal, Herbert-Voss, Krueger,
  Henighan, Child, Ramesh, Ziegler, Wu, Winter, Hesse, Chen, Sigler, Litwin,
  Gray, Chess, Clark, Berner, McCandlish, Radford, Sutskever, and
  Amodei}]{gpt3}
Tom Brown, Benjamin Mann, Nick Ryder, Melanie Subbiah, Jared~D Kaplan, Prafulla
  Dhariwal, Arvind Neelakantan, Pranav Shyam, Girish Sastry, Amanda Askell,
  Sandhini Agarwal, Ariel Herbert-Voss, Gretchen Krueger, Tom Henighan, Rewon
  Child, Aditya Ramesh, Daniel Ziegler, Jeffrey Wu, Clemens Winter, Chris
  Hesse, Mark Chen, Eric Sigler, Mateusz Litwin, Scott Gray, Benjamin Chess,
  Jack Clark, Christopher Berner, Sam McCandlish, Alec Radford, Ilya Sutskever,
  and Dario Amodei. 2020{\natexlab{a}}.
\newblock \href
  {https://proceedings.neurips.cc/paper/2020/file/1457c0d6bfcb4967418bfb8ac142f64a-Paper.pdf}
  {{Language Models are Few-Shot Learners}}.
\newblock In \emph{Advances in NIPS}, volume~33, pages 1877--1901. Curran
  Associates, Inc.

\bibitem[{Brown et~al.(2020{\natexlab{b}})Brown, Mann, Ryder, Subbiah, Kaplan,
  Dhariwal, Neelakantan, Shyam, Sastry, Askell, Agarwal, Herbert-Voss, Krueger,
  Henighan, Child, Ramesh, Ziegler, Wu, Winter, Hesse, Chen, Sigler, Litwin,
  Gray, Chess, Clark, Berner, McCandlish, Radford, Sutskever, and
  Amodei}]{NEURIPS2020_brown_gpt3}
Tom Brown, Benjamin Mann, Nick Ryder, Melanie Subbiah, Jared~D Kaplan, Prafulla
  Dhariwal, Arvind Neelakantan, Pranav Shyam, Girish Sastry, Amanda Askell,
  Sandhini Agarwal, Ariel Herbert-Voss, Gretchen Krueger, Tom Henighan, Rewon
  Child, Aditya Ramesh, Daniel Ziegler, Jeffrey Wu, Clemens Winter, Chris
  Hesse, Mark Chen, Eric Sigler, Mateusz Litwin, Scott Gray, Benjamin Chess,
  Jack Clark, Christopher Berner, Sam McCandlish, Alec Radford, Ilya Sutskever,
  and Dario Amodei. 2020{\natexlab{b}}.
\newblock \href
  {https://proceedings.neurips.cc/paper/2020/file/1457c0d6bfcb4967418bfb8ac142f64a-Paper.pdf}
  {Language models are few-shot learners}.
\newblock In \emph{Advances in Neural Information Processing Systems},
  volume~33, pages 1877--1901. Curran Associates, Inc.

\bibitem[{Brychc{\'\i}n and Habernal(2013)}]{Habernal.Brychcin.2013b}
Tom{\'a}{\v{s}} Brychc{\'\i}n and Ivan Habernal. 2013.
\newblock \href {https://aclanthology.org/R13-1016} {{Unsupervised Improving of
  Sentiment Analysis Using Global Target Context}}.
\newblock In \emph{Proceedings of the International Conference Recent Advances
  in Natural Language Processing {RANLP} 2013}, pages 122--128, Hissar,
  Bulgaria. INCOMA Ltd. Shoumen, BULGARIA.

\bibitem[{Chandra et~al.(2021)Chandra, Fahrizain, {Ibrahim}, and
  Laufried}]{chandra_survey_nonenglish_QA}
Andreas Chandra, Affandy Fahrizain, {Ibrahim}, and Simon~Willyanto Laufried.
  2021.
\newblock \href {https://doi.org/10.48550/ARXIV.2112.13634} {{A Survey on
  non-English Question Answering Dataset}}.

\bibitem[{{Chung} et~al.(2022){Chung}, {Hou}, {Longpre}, {Zoph}, {Tay},
  {Fedus}, {Li}, {Wang}, {Dehghani}, {Brahma}, {Webson}, {Gu}, {Dai}, {Suzgun},
  {Chen}, {Chowdhery}, {Castro-Ros}, {Pellat}, {Robinson}, {Valter}, {Narang},
  {Mishra}, {Yu}, {Zhao}, {Huang}, {Dai}, {Yu}, {Petrov}, {Chi}, {Dean},
  {Devlin}, {Roberts}, {Zhou}, {Le}, and {Wei}}]{chung2022_flan}
Hyung~Won {Chung}, Le~{Hou}, Shayne {Longpre}, Barret {Zoph}, Yi~{Tay}, William
  {Fedus}, Yunxuan {Li}, Xuezhi {Wang}, Mostafa {Dehghani}, Siddhartha
  {Brahma}, Albert {Webson}, Shixiang~Shane {Gu}, Zhuyun {Dai}, Mirac {Suzgun},
  Xinyun {Chen}, Aakanksha {Chowdhery}, Alex {Castro-Ros}, Marie {Pellat},
  Kevin {Robinson}, Dasha {Valter}, Sharan {Narang}, Gaurav {Mishra}, Adams
  {Yu}, Vincent {Zhao}, Yanping {Huang}, Andrew {Dai}, Hongkun {Yu}, Slav
  {Petrov}, Ed~H. {Chi}, Jeff {Dean}, Jacob {Devlin}, Adam {Roberts}, Denny
  {Zhou}, Quoc~V. {Le}, and Jason {Wei}. 2022.
\newblock \href {https://doi.org/10.48550/arXiv.2210.11416} {{Scaling
  Instruction-Finetuned Language Models}}.
\newblock \emph{arXiv e-prints}, page arXiv:2210.11416.

\bibitem[{Conneau et~al.(2018)Conneau, Rinott, Lample, Williams, Bowman,
  Schwenk, and Stoyanov}]{conneau2018xnli}
Alexis Conneau, Ruty Rinott, Guillaume Lample, Adina Williams, Samuel Bowman,
  Holger Schwenk, and Veselin Stoyanov. 2018.
\newblock \href {https://doi.org/10.18653/v1/D18-1269} {{XNLI: Evaluating
  Cross-lingual Sentence Representations}}.
\newblock In \emph{Proc. of the 2018 Conference on Empirical Methods in Natural
  Language Processing}, pages 2475--2485, Brussels, Belgium. ACL.

\bibitem[{Efimov et~al.(2019)Efimov, Boytsov, and
  Braslavski}]{efimov2019_sberquad}
Pavel Efimov, Leonid Boytsov, and Pavel Braslavski. 2019.
\newblock \href {http://arxiv.org/abs/1912.09723} {{SberQuAD -- Russian Reading
  Comprehension Dataset: Description and Analysis}}.
\newblock \emph{CoRR}, abs/1912.09723.

\bibitem[{Gao et~al.(2022)Gao, Madaan, Zhou, Alon, Liu, Yang, Callan, and
  Neubig}]{gao2022_pal}
Luyu Gao, Aman Madaan, Shuyan Zhou, Uri Alon, Pengfei Liu, Yiming Yang, Jamie
  Callan, and Graham Neubig. 2022.
\newblock \href {https://doi.org/10.48550/ARXIV.2211.10435} {{PAL:
  Program-aided Language Models}}.

\bibitem[{Koco{\'n} et~al.(2019)Koco{\'n}, Mi{\l}kowski, and
  Za{\'s}ko-Zieli{\'n}ska}]{kocon-etal-2019-multi}
Jan Koco{\'n}, Piotr Mi{\l}kowski, and Monika Za{\'s}ko-Zieli{\'n}ska. 2019.
\newblock \href {https://doi.org/10.18653/v1/K19-1092} {Multi-level sentiment
  analysis of {P}ol{E}mo 2.0: Extended corpus of multi-domain consumer
  reviews}.
\newblock In \emph{Proc. of the 23rd Conference on Computational Natural
  Language Learning (CoNLL)}, pages 980--991, Hong Kong, China. ACL.

\bibitem[{Lin(2004)}]{lin-2004-rouge}
Chin-Yew Lin. 2004.
\newblock \href {https://aclanthology.org/W04-1013} {{ROUGE: A Package for
  Automatic Evaluation of Summaries}}.
\newblock In \emph{Text Summarization Branches Out}, pages 74--81, Barcelona,
  Spain. ACL.

\bibitem[{Lu et~al.(2022)Lu, Bartolo, Moore, Riedel, and
  Stenetorp}]{lu-etal-2022-fantastically}
Yao Lu, Max Bartolo, Alastair Moore, Sebastian Riedel, and Pontus Stenetorp.
  2022.
\newblock \href {https://doi.org/10.18653/v1/2022.acl-long.556} {Fantastically
  ordered prompts and where to find them: Overcoming few-shot prompt order
  sensitivity}.
\newblock In \emph{Proceedings of the 60th Annual Meeting of the Association
  for Computational Linguistics (Volume 1: Long Papers)}, pages 8086--8098,
  Dublin, Ireland. Association for Computational Linguistics.

\bibitem[{Medve{\v d}(2022)}]{medved2022_sqad}
Marek Medve{\v d}. 2022.
\newblock \href {https://hdl.handle.net/11234/1-5019} {{SQAD} 3.2}.
\newblock {LINDAT}/\hskip0pt{CLARIAH}-{CZ} digital library at the Institute of
  Formal and Applied Linguistics ({{\'U}FAL}), Faculty of Mathematics and
  Physics, Charles University.

\bibitem[{Min et~al.(2022)Min, Lewis, Zettlemoyer, and
  Hajishirzi}]{min-etal-2022-metaicl}
Sewon Min, Mike Lewis, Luke Zettlemoyer, and Hannaneh Hajishirzi. 2022.
\newblock \href {https://doi.org/10.18653/v1/2022.naacl-main.201} {{M}eta{ICL}:
  Learning to learn in context}.
\newblock In \emph{Proceedings of the 2022 Conference of the North American
  Chapter of the Association for Computational Linguistics: Human Language
  Technologies}, pages 2791--2809, Seattle, United States. Association for
  Computational Linguistics.

\bibitem[{Mishra et~al.(2022)Mishra, Khashabi, Baral, and
  Hajishirzi}]{mishra-etal-2022-cross}
Swaroop Mishra, Daniel Khashabi, Chitta Baral, and Hannaneh Hajishirzi. 2022.
\newblock \href {https://doi.org/10.18653/v1/2022.acl-long.244} {Cross-task
  generalization via natural language crowdsourcing instructions}.
\newblock In \emph{Proceedings of the 60th Annual Meeting of the Association
  for Computational Linguistics (Volume 1: Long Papers)}, pages 3470--3487,
  Dublin, Ireland. Association for Computational Linguistics.

\bibitem[{Radford and Narasimhan(2018)}]{Radford2018gpt}
Alec Radford and Karthik Narasimhan. 2018.
\newblock \href
  {https://cdn.openai.com/research-covers/language-unsupervised/language_understanding_paper.pdf}
  {{Improving Language Understanding by Generative Pre-Training}}.

\bibitem[{Raffel et~al.(2020)Raffel, Shazeer, Roberts, Lee, Narang, Matena,
  Zhou, Li, and Liu}]{t5}
Colin Raffel, Noam Shazeer, Adam Roberts, Katherine Lee, Sharan Narang, Michael
  Matena, Yanqi Zhou, Wei Li, and Peter~J. Liu. 2020.
\newblock \href {http://jmlr.org/papers/v21/20-074.html} {Exploring the limits
  of transfer learning with a unified text-to-text transformer}.
\newblock \emph{Journal of Machine Learning Research}, 21(146):1--67.

\bibitem[{Rajpurkar et~al.(2016)Rajpurkar, Zhang, Lopyrev, and
  Liang}]{rajpurkar-etal-2016-squad}
Pranav Rajpurkar, Jian Zhang, Konstantin Lopyrev, and Percy Liang. 2016.
\newblock \href {https://doi.org/10.18653/v1/D16-1264} {{SQuAD: 100,000+
  Questions for Machine Comprehension of Text}}.
\newblock In \emph{Proc. of the 2016 Conference on Empirical Methods in Natural
  Language Processing}, pages 2383--2392, Austin, USA. ACL.

\bibitem[{Sanh et~al.(2022)Sanh, Webson, Raffel, Bach, Sutawika, Alyafeai,
  Chaffin, Stiegler, Raja, Dey, Bari, Xu, Thakker, Sharma, Szczechla, Kim,
  Chhablani, Nayak, Datta, Chang, Jiang, Wang, Manica, Shen, Yong, Pandey,
  Bawden, Wang, Neeraj, Rozen, Sharma, Santilli, Fevry, Fries, Teehan, Scao,
  Biderman, Gao, Wolf, and Rush}]{sanh2022multitask}
Victor Sanh, Albert Webson, Colin Raffel, Stephen Bach, Lintang Sutawika, Zaid
  Alyafeai, Antoine Chaffin, Arnaud Stiegler, Arun Raja, Manan Dey, M~Saiful
  Bari, Canwen Xu, Urmish Thakker, Shanya~Sharma Sharma, Eliza Szczechla,
  Taewoon Kim, Gunjan Chhablani, Nihal Nayak, Debajyoti Datta, Jonathan Chang,
  Mike Tian-Jian Jiang, Han Wang, Matteo Manica, Sheng Shen, Zheng~Xin Yong,
  Harshit Pandey, Rachel Bawden, Thomas Wang, Trishala Neeraj, Jos Rozen,
  Abheesht Sharma, Andrea Santilli, Thibault Fevry, Jason~Alan Fries, Ryan
  Teehan, Teven~Le Scao, Stella Biderman, Leo Gao, Thomas Wolf, and Alexander~M
  Rush. 2022.
\newblock \href {https://openreview.net/forum?id=9Vrb9D0WI4} {Multitask
  prompted training enables zero-shot task generalization}.
\newblock In \emph{International Conference on Learning Representations}.

\bibitem[{Sboev et~al.(2021)Sboev, Naumov, and Rybka}]{sboev2021data}
Alexander Sboev, Aleksandr Naumov, and Roman Rybka. 2021.
\newblock \href {https://doi.org/https://doi.org/10.1016/j.procs.2021.06.075}
  {{Data-Driven Model for Emotion Detection in Russian Texts}}.
\newblock \emph{Procedia Computer Science}, 190:637--642.
\newblock 2020 Annual International Conference on Brain-Inspired Cognitive
  Architectures for Artificial Intelligence: Eleventh Annual Meeting of the
  BICA Society.

\bibitem[{Scao et~al.(2022)Scao, Fan, Akiki, Pavlick, Ilić, Hesslow,
  Castagné, and Workshop{:}}]{scao2022_bloom}
Teven~Le Scao, Angela Fan, Christopher Akiki, Ellie Pavlick, Suzana Ilić,
  Daniel Hesslow, Roman Castagné, and BigScience. Workshop{:}. 2022.
\newblock \href {https://doi.org/10.48550/ARXIV.2211.05100} {{BLOOM: A
  176B-Parameter Open-Access Multilingual Language Model}}.

\bibitem[{{\v{S}}ev{\v{c}}{\'{\i}}kov{\'{a}}
  et~al.(2007){\v{S}}ev{\v{c}}{\'{\i}}kov{\'{a}}, {\v{Z}}abokrtsk{\'{y}}, and
  Kr{\r{u}}za}]{SevcikovaEtAl2007CNEC}
Magda {\v{S}}ev{\v{c}}{\'{\i}}kov{\'{a}}, Zden{\v{e}}k {\v{Z}}abokrtsk{\'{y}},
  and Old{\v{r}}ich Kr{\r{u}}za. 2007.
\newblock \href {https://doi.org/10.1007/978-3-540-74628-7_26} {{Named Entities
  in Czech: Annotating Data and Developing {NE} Tagger}}.
\newblock In \emph{{Proc. of the 10th International Conference on Text, Speech
  and Dialogue}}, volume 4629 of \emph{LNCS}, pages 188--195, Berlin /
  Heidelberg. Springer.

\bibitem[{Shavrina et~al.(2020)Shavrina, Fenogenova, Emelyanov, Shevelev,
  Artemova, Malykh, Mikhailov, Tikhonova, Chertok, and
  Evlampiev}]{shavrina2020russiansuperglue}
Tatiana Shavrina, Alena Fenogenova, Anton Emelyanov, Denis Shevelev, Ekaterina
  Artemova, Valentin Malykh, Vladislav Mikhailov, Maria Tikhonova, Andrey
  Chertok, and Andrey Evlampiev. 2020.
\newblock \href {https://doi.org/10.18653/v1/2020.emnlp-main.381}
  {{R}ussian{S}uper{GLUE}: A {R}ussian language understanding evaluation
  benchmark}.
\newblock In \emph{Proc. of the 2020 Conference on Empirical Methods in Natural
  Language Processing (EMNLP)}, pages 4717--4726. ACL.

\bibitem[{{\v S}tef{\'a}nik and Kadl{\v
  c}{\'{\i}}k(2022)}]{stefanik2022_not_in_context}
Michal {\v S}tef{\'a}nik and Marek Kadl{\v c}{\'{\i}}k. 2022.
\newblock \href {https://doi.org/10.48550/ARXIV.2212.01692} {{What is Not in
  the Context? Evaluation of Few-shot Learners with Informative
  Demonstrations}}.

\bibitem[{{\v{S}}tef{\'a}nik et~al.(2022){\v{S}}tef{\'a}nik, Novotn{\'y},
  Groverov{\'a}, and Sojka}]{stefanik-etal-2022-adaptor}
Michal {\v{S}}tef{\'a}nik, V{\'\i}t Novotn{\'y}, Nikola Groverov{\'a}, and Petr
  Sojka. 2022.
\newblock \href {https://doi.org/10.18653/v1/2022.acl-demo.26} {{Adaptor:
  Objective-Centric Adaptation Framework for Language Models}}.
\newblock In \emph{Proceedings of the 60th Annual Meeting of the ACL: System
  Demonstrations}, pages 261--269, Dublin, Ireland. ACL.

\bibitem[{Ullrich et~al.(2022)Ullrich, Drchal, Rýpar, Vincourová, and
  Moravec}]{ullrich2022_ctkfever}
Herbert Ullrich, Jan Drchal, Martin Rýpar, Hana Vincourová, and Václav
  Moravec. 2022.
\newblock \href {https://doi.org/10.48550/ARXIV.2201.11115} {{CsFEVER and
  CTKFacts: Acquiring Czech data for fact verification}}.

\bibitem[{Wang et~al.(2022)Wang, Mishra, Alipoormolabashi, Kordi, Mirzaei,
  Naik, Ashok, Dhanasekaran, Arunkumar, Stap, Pathak, Karamanolakis, Lai,
  Purohit, Mondal, Anderson, Kuznia, Doshi, Pal, Patel, Moradshahi, Parmar,
  Purohit, Varshney, Kaza, Verma, Puri, Karia, Doshi, Sampat, Mishra, Reddy~A,
  Patro, Dixit, and Shen}]{wang-etal-2022-super}
Yizhong Wang, Swaroop Mishra, Pegah Alipoormolabashi, Yeganeh Kordi, Amirreza
  Mirzaei, Atharva Naik, Arjun Ashok, Arut~Selvan Dhanasekaran, Anjana
  Arunkumar, David Stap, Eshaan Pathak, Giannis Karamanolakis, Haizhi Lai,
  Ishan Purohit, Ishani Mondal, Jacob Anderson, Kirby Kuznia, Krima Doshi,
  Kuntal~Kumar Pal, Maitreya Patel, Mehrad Moradshahi, Mihir Parmar, Mirali
  Purohit, Neeraj Varshney, Phani~Rohitha Kaza, Pulkit Verma, Ravsehaj~Singh
  Puri, Rushang Karia, Savan Doshi, Shailaja~Keyur Sampat, Siddhartha Mishra,
  Sujan Reddy~A, Sumanta Patro, Tanay Dixit, and Xudong Shen. 2022.
\newblock \href {https://aclanthology.org/2022.emnlp-main.340}
  {Super-{N}atural{I}nstructions: Generalization via declarative instructions
  on 1600+ {NLP} tasks}.
\newblock In \emph{Proceedings of the 2022 Conference on Empirical Methods in
  Natural Language Processing}, pages 5085--5109, Abu Dhabi, United Arab
  Emirates. Association for Computational Linguistics.

\bibitem[{Wei et~al.(2022)Wei, Bosma, Zhao, Guu, Yu, Lester, Du, Dai, and
  Le}]{wei2022finetuned}
Jason Wei, Maarten Bosma, Vincent Zhao, Kelvin Guu, Adams~Wei Yu, Brian Lester,
  Nan Du, Andrew~M. Dai, and Quoc~V Le. 2022.
\newblock \href {https://openreview.net/forum?id=gEZrGCozdqR} {{Finetuned
  Language Models are Zero-Shot Learners}}.
\newblock In \emph{Proc. of International Conference on Learning
  Representations}.

\bibitem[{Wolf et~al.(2020)Wolf, Debut, Sanh, Chaumond, Delangue, Moi, Cistac,
  Rault, Louf, Funtowicz, Davison, Shleifer, von Platen, Ma, Jernite, Plu, Xu,
  Le~Scao, Gugger, Drame, Lhoest, and Rush}]{Wolf2019HuggingFacesTS}
Thomas Wolf, Lysandre Debut, Victor Sanh, Julien Chaumond, Clement Delangue,
  Anthony Moi, Pierric Cistac, Tim Rault, Remi Louf, Morgan Funtowicz, Joe
  Davison, Sam Shleifer, Patrick von Platen, Clara Ma, Yacine Jernite, Julien
  Plu, Canwen Xu, Teven Le~Scao, Sylvain Gugger, Mariama Drame, Quentin Lhoest,
  and Alexander Rush. 2020.
\newblock \href {https://doi.org/10.18653/v1/2020.emnlp-demos.6}
  {{Transformers: State-of-the-Art Natural Language Processing}}.
\newblock In \emph{Proc. of the 2020 Conf. EMNLP: System Demonstrations}, pages
  38--45. ACL.

\bibitem[{Wr{\'o}blewska et~al.(2017)Wr{\'o}blewska, Krasnowska-Kiera{\'s}, and
  Katarzyna}]{wroblewska-krasnowska-kieras-2017-polish}
Alina Wr{\'o}blewska, Krasnowska-Kiera{\'s}, and Katarzyna. 2017.
\newblock \href {https://doi.org/10.18653/v1/P17-1073} {{P}olish evaluation
  dataset for compositional distributional semantics models}.
\newblock In \emph{Proc. of the 55th Annual Meeting of the ACL (Volume 1: Long
  Papers)}, pages 784--792, Vancouver, Canada. ACL.

\bibitem[{Xue et~al.(2021)Xue, Constant, Roberts, Kale, Al-Rfou, Siddhant,
  Barua, and Raffel}]{xue-etal-2021-mt5}
Linting Xue, Noah Constant, Adam Roberts, Mihir Kale, Rami Al-Rfou, Aditya
  Siddhant, Aditya Barua, and Colin Raffel. 2021.
\newblock \href {https://doi.org/10.18653/v1/2021.naacl-main.41} {m{T}5: A
  massively multilingual pre-trained text-to-text transformer}.
\newblock In \emph{Proceedings of the 2021 Conference of the North American
  Chapter of the Association for Computational Linguistics: Human Language
  Technologies}, pages 483--498, Online. Association for Computational
  Linguistics.

\bibitem[{Zhang et~al.(2022)Zhang, Roller, Goyal, Artetxe, Chen, Chen, Dewan,
  Diab, Li, Lin, Mihaylov, Ott, Shleifer, Shuster, Simig, Koura, Sridhar, Wang,
  and Zettlemoyer}]{zhang2022opt}
Susan Zhang, Stephen Roller, Naman Goyal, Mikel Artetxe, Moya Chen, Shuohui
  Chen, Christopher Dewan, Mona Diab, Xian Li, Xi~Victoria Lin, Todor Mihaylov,
  Myle Ott, Sam Shleifer, Kurt Shuster, Daniel Simig, Punit~Singh Koura, Anjali
  Sridhar, Tianlu Wang, and Luke Zettlemoyer. 2022.
\newblock \href {http://arxiv.org/abs/2205.01068} {Opt: Open pre-trained
  transformer language models}.

\end{thebibliography}
%\printbibliography

\appendix

\section{Table of templates}
\label{appx:templates}

Table~\ref{tab:all_templates} contains a full list of templates collected within this work, including the segments filled from the transformed datasets.

% toy project: https://www.overleaf.com/2462483728fphtvzcvmgdp
\def\RUS#1{\fontfamily{qtm}\fontencoding{T2A}\selectfont#1}
\begin{table*}[p]
\enlargethispage*{3\baselineskip}\renewcommand{\baselinestretch}{0.94}
 \centerline{\scalebox{.77}{%
 \begin{tabular}{@{}l@{\,}l@{\ \ }p{1.2\textwidth}@{}}
 \toprule
 Lang & Task & Template\\
\midrule
 cs       & NER  & \{\{text\}\} \{\{label\_type\}\} v tomto textu je                                                                                                  \\
 cs       & NER  & Jaká entita typu \{\{label\_type\}\} se nachází v následujícím textu? \{\{text\}\}                                              \\
 cs       & NER  & \{\{text\}\} Jaká entita typu \{\{label\_type\}\} se nachází v předchozím odstavci?                                                                                   \\
 cs       & Clf. & Jaký sentiment vyjadřuje následující filmová recenze? \{\{comment\}\}                                                                                                 \\
 cs       & Clf. & \{\{comment\}\} Shledal recenzent tento film \{\{"dobrým nebo zlým"\}\}?                                                                                              \\
 cs       & Clf. & \{\{comment\}\}  Je tato recenze \{\{"pozitivní nebo negativní"\}\}?                                                                                                  \\
 cs       & Clf. & \{\{comment\}\} Je tento komentář \{\{"pozitivní, neutrální nebo negativní"\}\}?                                                                                      \\
 cs       & Clf. & \{\{comment\}\} Jaký je sentiment tohoto komentáře? \{\{"pozitivní, neutrální nebo negativní"\}\}?                                                                    \\
 cs       & Clf. & Jaký sentiment má následující komentář? \{\{comment\}\}                                                                                                               \\
 cs       & Clf. & \{\{comment\}\} Je tato recenze \{\{"pozitivní, neutrální nebo negativní"\}\}?                                                                                        \\
 cs       & Clf. & Jaký sentiment má následující recenze? \{\{comment\}\}                                                                                                                \\
 cs       & Clf. & \{\{comment\}\} Jaký je sentiment této recenze? \{\{"pozitivní, neutrální nebo negativní"\}\}?                                                                        \\
 cs       & QA   & \{\{context\}\}  Q: \{\{question\}\}  S odkazem na sekci výše je správná odpověď na danou otázku                                                                      \\
 cs       & QA   & Podívejte se na odstavec níže a odpovězte na následující otázku:  Odstavec: \{\{context\}\}  Otázka: \{\{question\}\}                                                 \\
 cs       & QA   & \{\{context\}\}  S odkazem na výše uvedený odstavec, \{\{question\}\}                                                                                                 \\
 cs       & QA   & \{\{context\}\}  Otázka: \{\{question\}\}  Odpověď:                                         \\
 cs       & NLI  & \{\{evidence\}\} Otázka: \{\{claim\}\} Pravda, nepravda, nebo ani jedno?                                                  \\
 cs       & NLI  & \{\{evidence\}\} Za uvedeného předpokladu a na základě znalostí o světe, "\{\{claim\}\}" je určitě pravda, nepravda, nebo není jasné?                                 \\
 cs       & NLI  & \{\{evidence\}\} Na základě předchozího odstavce, je to pravda, že "\{\{claim\}\}"? Ne, možná, nebo ano?                                                                \\
 cs       & NLI  & Za předpokladu, že \{\{evidence\}\} vyplývá, že \{\{claim\}\}? Ano, ne, nebo možná?                                                                                   \\
 cs       & NLI  & Předpokládejme následovné: \{\{evidence\}\} Pak musí být pravda, že "\{\{claim\}\}"? Ano, ne, nebo možná?                                                             \\
 cs       & NLI  & Předpokládáme, že \{\{evidence\}\} Je možné předpokládat, že "\{\{claim\}\}" je pravda? Ano, ne, nebo možná?                                                          \\
 cs       & NLI  & Předpokládejme následovné: \{\{evidence\}\} Pak následující tvrzení: "\{\{claim\}\}" je pravda, nepravda, nebo nejasné?                                               \\
 \addlinespace
 pl       & clf. & "\{\{text\}\}" Ten tekst jest pozytywny, negatywny, neutralny czy dwuznaczny?                                                                                         \\
 pl       & clf. & Oceń ten tekst jako pozytywny, negatywny, neutralny lub dwuznaczny.   Tekst: \{\{text\}\}                                                                             \\
 pl       & clf. & Oceń wydźwięk tego tekstu jako pozytywny, negatywny, neutralny lub dwuznaczny.   Tekst: \{\{text\}\}   Wydźwięk:                                                      \\
 pl       & clf. & "\{\{text\}\}" Jaka jest ta recenzja? Jest pozytywna, negatywna, neutralna czy dwuznaczna?:                                                                           \\
 pl       & NLI  & "\{\{sentence\_A\}\}" Na podstawie tego, można powiedzieć, że zdanie "\{\{sentence\_B\}\}" jest potwierdzeniem, zaprzeczeniem czy niezwiązane?                        \\
 pl       & NLI  & Oceń czy poniższe zdania są zgodne ze sobą - tak, nie czy nie wiadomo?   Zdanie A: \{\{sentence\_A\}\}  Zdanie B: \{\{sentence\_B\}\}   Zgodność:                     \\
 pl       & NLI  & Hipotezę i przesłankę można powiązać jako potwierdzenie, zaprzeczenie lub niezwiązane.   Hipoteza: \{\{sentence\_A\}\}  Przesłanka: \{\{sentence\_B\}\}   Powiązanie: \\
 pl       & NLI  & Hipoteza: \{\{sentence\_A\}\}  Przesłanka: \{\{sentence\_B\}\}  Czy przesłanka jest dla hipotezy potwierdzeniem, zaprzeczeniem czy jest niezwiązana?                  \\
 pl       & NER  & "\{\{text\}\}" \{\{label\_type\_selected\}\} w tym tekście to                                                                                                         \\
 pl       & NER  & Znajdź encje typu \{\{label\_type\_selected\}\} w następującym tekście: \{\{text\}\}                                                                                  \\
 pl       & NER  & Jaka encja typu \{\{label\_type\_selected\}\} znajduje się w następującym tekście? "\{\{text\}\}"                                                                     \\
 pl       & NER  & "\{\{text\}\}" Jaka encja typu \{\{label\_type\_selected\}\} znajduje się w poprzednim akapicie?                                                                      \\
 pl       & NER  & "\{\{text\}\}" \{\{label\_type\_selected\}\} w tym tekście to                                                                                                         \\
 pl       & NER  & Znajdź encje typu \{\{label\_type\_selected\}\} w następującym tekście: \{\{text\}\}                                                                                  \\
 pl       & NER  & Jaka encja typu \{\{label\_type\_selected\}\} znajduje się w następującym tekście? "\{\{text\}\}" \\
 pl       & NER  & "\{\{text\}\}" Jaka encja typu \{\{label\_type\_selected\}\} znajduje się w poprzednim akapicie? \\
 \addlinespace
 ru       & NER  & \{\{text\}\} \{\{label\_type\}\} \RUS{в этом тексте:} \\
 ru       & NER  & \RUS{Какой объект типа} \{\{label\_type\}\} \RUS{встречается в следующем тексте?} \{\{text\}\} \\
 ru       & NER  & \{\{text\}\} \RUS{Какой объект типа} \{\{label\_type\}\} \RUS{находится в предыдущем абзаце}? \\
 ru       & NLI  & \{\{premise\}\} \RUS{Используя только приведенное выше описание и то, что вы знаете о мир,} "\{\{hypothesis\}\}" \RUS{определенно верна, неверна или неубедительна?}              \\
 ru       & NLI  & \{\{premise\}\} \RUS{Верно ли, исходя из предыдущего отрывка, что} "\{\{hypothesis\}\}"? \RUS{Да, нет, а может быть}?                                                             \\
 ru       & NLI  & \RUS{Учитывая} \{\{premise\}\}, \RUS{следует ли из этого, что} "\{\{hypothesis\}\}"? \RUS{Да, нет или возможно}?                                                                        \\
 ru       & NLI  & \{\{premise\}\} \RUS{Имеем ли мы право говорить, что} "\{\{hypothesis\}\}"? \RUS{Да, нет, или может быть?}                                                                        \\
 ru       & NLI  & \RUS{Учитывая, что} \{\{premise\}\} \RUS{Следовательно, должно быть верно, что} "\{\{hypothesis\}\}"? \RUS{Да, нет, а Возможно}?                                                        \\
 ru       & NLI  & \RUS{Учитывая} \{\{premise\}\} \RUS{Должны ли мы предположить, что} "\{\{hypothesis\}\}" \RUS{верна? Да, нет или возможно}?                                                             \\
 ru       & NLI  & \RUS{Примите за истину следующее:} \{\{premise\}\} \RUS{Тогда следующее утверждение:} "\{\{hypothesis\}\}" \RUS{есть "правда", "ложь" или "неубедительно"}?                             \\
 ru       & QA   & \{\{context\}\} \RUS{Ответ на вопрос:} \{\{question\}\}                                                                                                                     \\
 ru       & QA   & \RUS{Посмотрите на абзац ниже и ответьте на следующий вопрос:  Абзац:} \{\{context\}\}  \RUS{Вопрос:} \{\{question\}\}                                                            \\
 ru       & QA   & \{\{context\}\}\textbackslash{}n\textbackslash{}n\RUS{Со ссылкой на абзац выше,} \{\{question\}\}                                                                           \\
 ru       & QA   & \{\{context\}\}  \RUS{Вопрос:} \{\{question\}\}  \RUS{Отвечать:}                                                                                                                  \\
 ru       & Clf. & \{\{text\}\} \RUS{Это обзор радят, печал, удивление, страх или гнев}?                                                                                                       \\
 ru       & Clf. & \RUS{Каково настроение следующего обзора?} \{\{text\}\} \RUS{Варианты: радость, печаль, удивление, страх, гнев}                                                                   \\
 ru       & Clf. & \{\{text\}\} \RUS{Каково настроение этого обзора? радость, печаль, удивление, страх или гнев?}                                                                              \\
 \bottomrule
 \end{tabular}}}
 \caption{Templates for all languages and all task types that we collect in this work. Templates were written by native speakers of the template's language.}
\label{tab:all_templates}
\end{table*}

\section{Details of training and evaluation configuration}
\label{appx:details}

All models trained within this work, including the baselines are based on the mT5-Large model trained on the referenced dataset(s) using Batch size=30, learning rate = $2\cdot10^-5$ and early stopping with the patience of 10~evaluations (i.e. 2,000 updates) based on the evaluation loss on a held-out set of data of all training datasets. Where the validation split was provided, we use it as the held-out evaluation set, otherwise, we slice out the last 200~samples of the training data for this purpose. For a simple tracking of multi-dataset training, as well as for convenient bulk training of all supervised baselines, we used Adaptor library~\cite{stefanik-etal-2022-adaptor} in version 0.2.0, with Hugging Face Transformers library \cite{Wolf2019HuggingFacesTS}, version 4.19.1 as backend. For each training, we used a single Nvidia A100 with 80\,GB of GPU memory.

In all evaluations, we used greedy search generation with a default configuration of \textit{generate} method in version 4.19.1.
\end{document}